\documentclass[journal]{IEEEtran}

\usepackage[utf8]{inputenc} 
\usepackage[T1]{fontenc}    
\usepackage{hyperref}       
\usepackage{url}            
\usepackage{booktabs}       
\usepackage{amsfonts}       
\usepackage{nicefrac}       
\usepackage{microtype}      
\usepackage{lipsum}
\usepackage{fancyhdr}       
\usepackage{graphicx}       
\usepackage{amsmath}
\usepackage{algpseudocode}
\usepackage{algorithm}
\usepackage{array}
\usepackage{float}
\usepackage{wrapfig}
\usepackage{multirow}
\usepackage[symbol]{footmisc}
\usepackage{tablefootnote}
\usepackage{xcolor}
\usepackage{breqn}
\usepackage{mathtools, cuted}

\usepackage{caption,setspace}
\graphicspath{{images/}}     

\algnewcommand\algorithmicforeach{\textbf{for each}}
\algdef{S}[FOR]{ForEach}[1]{\algorithmicforeach\ #1\ \algorithmicdo}

\newcommand{\longcomment}[1]{}

\DeclareRobustCommand*{\IEEEauthorrefmark}[1]{%
  \raisebox{0pt}[0pt][0pt]{\textsuperscript{\footnotesize #1}}%
}

\newif\ifreview
\reviewfalse

\newcommand{\revised}[1]{%
    \ifreview
        {\color{blue}#1}%
    \else 
        #1%
    \fi  
}
  
\begin{document}

\frenchspacing

\title{Combining Constructive and Perturbative Deep Learning Algorithms for \revised{the Capacitated Vehicle Routing Problem}}
\author{
\IEEEauthorblockN{Roberto García-Torres\IEEEauthorrefmark{1}, Alitzel Adriana Macias-Infante\IEEEauthorrefmark{2} Santiago Enrique~Conant-Pablos\IEEEauthorrefmark{3} \\ Jos\'{e} Carlos Ortiz-Bayliss\IEEEauthorrefmark{4} Hugo Terashima-Mar\'{i}n\IEEEauthorrefmark{5}}
\vspace{0.07in}
\IEEEauthorblockA{\\School of Engineering and Sciences, Tecnologico de Monterrey.}
\IEEEauthorblockA{\\E-mails: \IEEEauthorrefmark{1}\emph{garcia.roberto@exatec.tec.mx}, \IEEEauthorrefmark{2}\emph{A01373166@exatec.tec.mx}, \IEEEauthorrefmark{3}\emph{sconant@tec.mx}, \\\IEEEauthorrefmark{4}\emph{jcobayliss@tec.mx}, \IEEEauthorrefmark{5}\emph{terashima@tec.mx}}
}

\markboth
{García-Torres \MakeLowercase{\textit{et al.}} Combining Constructive and Perturbative Deep Learning Algorithms for CVRP}
{García-Torres \MakeLowercase{\textit{et al.}}: Combining Constructive and Perturbative Deep Learning Algorithms for CVRP}
\maketitle

\begin{abstract}

\revised{The Capacitated Vehicle Routing Problem is a well-known NP-hard problem that poses the challenge of finding the optimal route of a vehicle delivering products to multiple locations. Recently, new efforts have emerged to create constructive and perturbative heuristics to tackle this problem using Deep Learning. In this paper, we join these efforts to develop the Combined Deep Constructor and Perturbator, which combines two powerful constructive and perturbative Deep Learning-based heuristics, using attention mechanisms at their core. Furthermore, we improve the Attention Model-Dynamic for the Capacitated Vehicle Routing Problem by proposing a memory-efficient algorithm that reduces its memory complexity by a factor of the number of nodes. Our method shows promising results. It demonstrates a cost improvement in common datasets when compared against other multiple Deep Learning methods. It also obtains close results to the state-of-the art heuristics from the Operations Research field. Additionally, the proposed memory efficient algorithm for the Attention Model-Dynamic model enables its use in problem instances with more than 100 nodes.}
\end{abstract}


\section{Introduction}
\label{sec:Introduction}

This work addresses the Capacitated Vehicle Routing Problem~(CVRP)~\cite{VRP} for a single vehicle. This problem involves optimizing the route cost of a vehicle delivering products to several locations. Each location has a demand for the product the vehicle is delivering. In addition, such a vehicle has a maximum carrying capacity of the product and can make refills every time it returns to a particular location, denoted as the depot. The latest research proposes two scopes to address the CVRP and some of its variants\revised{: the Operations Research perspective and the Machine Learning one, the last being the most recent}. \revised{Regarding Operations Research}, the most popular heuristics are the Lin-Kernighan-Helsgaun~(LKH)~\cite{HELSGAUN2000106}, mainly its 3\revised{\textsuperscript{rd}} version \revised{(}LKH3\revised{)}, and the more recent Hybrid Genetic Search~(HGS) for the CVRP~\cite{DBLP:journals/corr/abs-2012-10384, hgs}. LKH3 is commonly used as a baseline comparison. However, \revised{HGS} has recently shown state-of-the-art results in the \revised{Operations Research} field. Besides, there has been a rapid growth in the quality of the proposed models in the \revised{Machine Learning} domain. Popular approaches are the Pointer Networks~\cite{NIPS2015_29921001}, the Attention Model-Dynamic~(AM-D) first proposed by Kool et al.~\cite{kool2019attention},  Deep Policy Dynamic Programming~\cite{kool2021deep}, and the outstanding algorithm proposed by Lu et al.~\cite{Lu2020A}, which currently holds the state-of-the-art results in the Machine Learning field.

The main focus of this research relates to the Machine Learning perspective. As we have found, recent Machine Learning methods in the CVRP domain are divided into two main ways of obtaining a solution: constructive and perturbative heuristics. Constructive heuristics start from an empty solution and build the solution step by step. Commonly, they use a neural network policy that encodes the graph structure of the CVRP and selects the next node to add to the solution path. It repeats this process until all the nodes are visited. Conversely, perturbative heuristics start from an already created solution and proceed to apply \texttt{destroy} and \texttt{repair} operators to it. Frequently, they use a neural network that encodes the graph structure of the CVRP along with its current solution. Besides, these methods use the obtained embeddings as input to a \revised{Neural Network} representing the \texttt{destroy} and \texttt{repair} operators. Such operators choose which nodes to destroy and repair in the current solution.

Usually, perturbative methods begin with a solution generated by a naive constructive heuristic. An example is the model proposed by Gao et al.~\cite{gao2020learn}, which we refer to as the a \revised{L}ocal-\revised{S}earch \revised{H}euristic \revised{(LSH)} in the rest of this paper. The LSH  generates initial solutions by randomly iterating over all the nodes in the CVRP and inserting them into its solution sequence~(initially empty) wherever they cause the smallest cost~(distance) increase to the solution at each iteration step. In this paper, we refer to this re-insertion process as the minimum cost principle. The solutions are represented by the sequence of nodes to be visited in the order in which they appear. Once solutions are generated for multiple CVRPs following this methodology, a model learns to apply \texttt{destroy} and \texttt{repair} operators to the solutions. In this paper, we explore the idea of using a far better than naive, Deep Learning\revised{-}based constructive algorithm to generate initial solutions for a \revised{Deep Learning-based} perturbative algorithm. We are driven by techniques like this one showing success in popular non \revised{Deep Learning} algorithms such as LKH.

This work proposes a model that combines outstanding \revised{Deep Learning}-based constructive and perturbative algorithms. For the constructor, we use the AM-D, initially proposed by Kool et al.~\cite{kool2019attention} and further improved by Peng et al.~\cite{peng2020deep}. We use the LSH proposed by Gao et al.~\cite{gao2020learn} for the perturbator. Furthermore, we propose an improvement to the training algorithm of the AM-D. This improvement reduces its memory complexity by a factor of the number of nodes while maintaining its runtime complexity in the same order. To demonstrate the effectiveness of the proposed method, we tested it in the CVRP data distribution \revised{suggested} by Nazari et al.~\cite{nazari2018reinforcement}. In this data distribution, CVRPs consist of the number of locations, the bi-dimensional coordinates of the locations, the demand of each location, and the vehicle's capacity. In it, the coordinates of each location follow a uniform distribution bounded by 0 and 1 and the demands follow a discrete uniform distribution bounded by 1 and 9. The number of nodes and the vehicle's capacity \revised{are} predefined (and the same) for all CVRPs. Experiment results \revised{are favorable. The} Combined Deep Constructor and Perturbator (CDCP) outperforms \revised{multiple Deep Learning-based methods and achieves relatively close results to the state-of-the-art heuristics from the Operations Research} domain. Similarly, we \revised{revamp} the training algorithm on the AM-D showing immense memory consumption improvements on the same data distribution. This enables using this constructive algorithm on CVRPs with more than 100 nodes without the need for vast amounts of GPU memory\revised{, as was the case in previous proposals,} in the order of tens of gigabytes.

The rest of this paper is organized as follows. We discuss the related work in Sect.~\ref{section:related_work}. We later describe the AM-D and the LSH in Sects.~\ref{section:constructive_algorithm} and \ref{section:perturbative_algorithm}, respectively. Further, we present our proposal in Sect.~\ref{section:proposal}. Section~\ref{section:experiments} presents the experiments and discusses the results obtained. Finally, the conclusion and future work is depicted in Sect.~\ref{section:conclusion}. 

\section{Related Work}
\label{section:related_work}

\revised{
Currently, we identify two main branches for solving combinatorial optimization problems such as the CVRP. The line dividing these branches is thin. However, we can roughly say the first one is through Machine Learning and Deep Learning algorithms and the second one is through heuristic and meta-heuristic methods. Some of the most popular general heuristics and meta-heuristics are simulated annealing~\cite{Kirkpatrick1983}, variable neighborhood search~\cite{MLADENOVIC19971097}, large neighborhood search~\cite{Pisinger2010}, amongst others which allow for escaping local optima in local search~\cite{Bai2021}. In the specific case of the CVRP, two of the leading state-of-the-art heuristics are LKH3~\cite{Helsgaun2017} and HGS~\cite{DBLP:journals/corr/abs-2012-10384}. Nevertheless, in this paper we focus on the Machine Learning and Deep Learning branch. In particular, we center in perturbative and constructive Deep Learning algorithms. In the related work we delve into these methodologies.

First, we discuss the constructive Deep Learning algorithms. These are characterized by sequentially constructing a solution that is a permutation of the input nodes. The Pointer Network was introduced by Vinyals et al.~\cite{NIPS2015_29921001} as one of the first models in this domain. In their work, they implement a neural attention mechanism for solving combinatorial optimization problems of variable input size. In concrete, their model sequentially generates a permutation of the input which serves as the problem’s solution sequence. Their approach was demonstrated to be effective in solving the Traveling Salesman Problem, a variant of the Vehicle Routing Problem. An aspect to consider about this model is that it uses supervised learning. This property introduces some side effects related to the optimality of the solutions and scalability of training.

Following the work done with Pointer Networks, Nazari et al. present a simplified version of this model. In their work, they introduce a single-policy model that constructs a solution as a sequence of consecutive actions~\cite{nazari2018reinforcement}. However, different from the Pointer Network, their model is trained using Reinforcement Learning. Particularly, their approach uses a Recurrent Neural Network decoder coupled with an attention mechanism trained using an actor-critic policy gradient algorithm. They also propose the CVRP data distribution that subsequent works, including ours, use for evaluation and comparison.

Kool et al. present a graph attention network and a model based on the Pointer Network~\cite{kool2019attention}. As in the Pointer Network, they use attention mechanisms for solving the combinatorial optimization problem. Additionally, similar to Nazari’s et al. approach, they use Reinforcement Learning for training. Specifically, they train their model using the REINFORCE algorithm with a greedy rollout baseline. Nevertheless, the core of their work relies on the use of the transformer architecture. Their model outperforms the Pointer Network, both due to the attention model they implement and to the rollout baseline. Following this line of work, Peng et al. also propose an attention model that constructively builds a solution for CVRPs~\cite{peng2020deep}. They propose a more dynamic version of the work of Kool et al.~\cite{kool2019attention}. Their approach relies on recomputing the embeddings of each node in the CVRP more frequently than the original implementation. This is the constructive algorithm we work with in this paper.

Other related works have been introduced recently. Sheng et al. propose a variation of Pointer Networks~\cite{Sheng2020}. Their model considers the actual distribution conditions of the input to relate input nodes and output decisions. They introduce a global mechanism that learns the aforementioned relationship. Additionally, Lu et al. explore the idea of combining reinforcement and supervised learning through a graph convolutional network with node and edge features~\cite{Lu2020}. Their proposal results in a model that accelerates convergence and improves solutions, which is useful for large-sized problems. 

Second, we discuss the perturbative algorithms. These are methods that iteratively improve an initial solution instead of constructively building one. Chen and Tian tackle this problem using deep Reinforcement Learning to improve an initial given solution. Their algorithm works by selecting a local region from which to remove a node and then a heuristic rule to insert that node back to the solution~\cite{chen2019learning}. Their approach does not use attention mechanisms, instead it resorts to a bi-directional Long Short Term Memory model to embed the routes. It works by learning a neural rule-picking policy using the Advantage Actor-Critic algorithm. Da Costa et al. implement a model that learns an improvement heuristic based on 2-opt operators~\cite{daCosta2021}. This model outputs an action per step, and it can easily be extended to $k$-opt operators. Their method takes advantage of attention mechanisms to output node sequences to action on. Additionally, the architecture they propose uses both graph convolutional layers and recurrent layers. It also leverages Reinforcement Learning by training using the policy gradient algorithm.

Veličković et al. introduce Graph Attention Networks~(GAT), which allow graph topologies to be encoded~\cite{gat2018}. This conclusion-style neural network uses self-attention layers on graph-structured data. Although the authors do not directly apply the GAT to the CVRP, a variation of this architecture is used in the perturbative algorithm we leverage in this paper.

Gao et al. propose a modified version of GAT, Element-wise Graph Attention Network (EGATE), which complements the node embeddings by adding information regarding the edges between nodes~\cite{gao2020learn}. They also propose an attention model that learns to design a universal heuristic. This heuristic recursively improves the solution of CVRPs by applying a perturbation that removes and then reinserts the nodes of the solution until it converges to a local optimum. The training leverages an actor-critic algorithm which learns the removal pattern and the reinsertion order. This is the perturbative model we work with in our proposal. Additionally, Chen et al. propose a similar model that relies on adaptive large neighborhood search ~\cite{Chen2020}. In their work, similar to the proposal from Gao et al. ~\cite{gao2020learn}, they define a \texttt{destroy} and \texttt{repair} operator. The difference is that Chen et al. present a \texttt{destroy} operator that is dynamically determined~\cite{Chen2020}. Their results reflect their model holds good awareness of spatial and context information.

Other approaches, different from perturbative and constructive algorithms which are not in the main domain of this paper, have been recently proposed. We will mention a few of them to provide some context. Kool et at. present a model that combines two different methods: dynamic programming algorithms and learned natural heuristics~\cite{kool2021deep}. They restrict the dynamic programming space through a deep neural network policy and search for a solution in this space. Kwon et al. introduce a data-driven combinatorial optimization approach that uses a modified REINFORCE algorithm~\cite{kwon2021pomo}. Their deep Reinforcement Learning model works by leveraging the existence of multiple optimal solutions. This is enabled by the use of symmetries found in the representation of the combinatorial optimization problem. 

As the constructive algorithm by Peng et al.~\cite{peng2020deep} and the perturbative model by Gao et al.~\cite{gao2020learn} are the main related works which our method works with, we will detail them further in the next sections.
}

\section{Constructive Algorithm}
\label{section:constructive_algorithm}

\revised{T}his work uses the AM-D as the constructive algorithm for providing the initial CVRP solutions. Initially proposed by Kool et al.~\cite{kool2019attention} and further improved by Peng et al.~\cite{peng2020deep}, the AM-D is a robust model that leverages \revised{Reinforcement Learning} with \revised{Deep Learning} using the transformers architecture. Trained using the REINFORCE with baseline algorithm~\cite{Williams1992}, this model procedurally constructs a solution for a given CVRP.

\subsection{Model Encoder}

The encoder seeks to capture the relationship among nodes in the graph. It starts by applying a linear transformation to the three-dimensional vector of each node, indicating its bi-dimensional position and demand. Next, we apply a Multi-head Attention~(MHA) mechanism~\cite{NIPS2017_3f5ee243} using the obtained embeddings as queries, keys and values. We stack various MHA mechanisms where the embeddings resulting from the output of the previous MHA layer ($MHA_{i-1}$) are the inputs to the following MHA layer~($MHA_{i}$). In practice, we also include a skip-connection mechanism between each MHA layer. Such a mechanism takes as input the embeddings computed at $MHA_{i}$ and $MHA_{i-1}$ and forwards its output as the input embedding for the next MHA layer $MHA_{i+1}$. Finally, we repeat this process multiple times until we end up with the final embeddings for each node at the last MHA layer.

\subsection{Model Decoder}

Once we have embeddings calculated for each node, we use the decoder to produce the next node to add to the solution sequence. The decoder uses an attention mechanism similar to the encoder’s one. However, here we use a single query vector which is calculated by concatenating:

\begin{enumerate}
  \item The mean of the node embeddings (representing the graph).
  \item The embedding of the previous node (initially, at $t=1$, the depot embedding).
  \item The current remaining capacity of the vehicle.
\end{enumerate}

Further, a projection of the generated query vector serves as the query input of an Attention Mechanism. In it, the keys and values are projections of the node embeddings we calculated with the decoder. Additionally, we incorporate a masking procedure in this Attention Mechanism that masks nodes which are already part of the solution (except for the depot). By doing so, \revised{the attention step ignores such} embeddings. The resulting embedding is the context vector representing the current state of the CVRP.

The context vector mentioned before serves as input to a final matrix product with a projection of the encoder embeddings. This is followed by a SoftMax layer resulting in a distribution that indicates the probability of each node being the next in the solution sequence. It is important to note that we also use a masking procedure in this step. Such a masking procedure masks the embeddings of nodes that form part of the solution already or nodes that are not reachable from the last selected node. The latter case may occur due to a lack of vehicle capacity to fulfill the node demands.

\subsection{Combining Encoder and Decoder}

To combine the encoder and decoder, first, we compute the embeddings of the CVRPs using the encoder. Then we feed those embeddings to the decoder. The decoder generates a new node at each step and adds it to the current tour (a tour is a path from the depot to the depot going through one or more locations). Finally, every time we finish a tour, that is, when the vehicle returns to the depot, we create new embeddings using the encoder. We repeat this process until there are no more locations to visit. It is important to note that we generate the node embeddings mentioned by masking out the nodes that are already part of the solution.

\subsection{Training Algorithm}

As mentioned, \revised{we conducted} the training using the REINFORCE with baseline algorithm~\cite{Williams1992}. In this algorithm, we define the gradient of the loss of the policy $p_{\theta}$, parametrized by $\theta$, with Eq.~\ref{eqn:1}.

\begin{dmath}
    \label{eqn:1}
    \nabla_{\theta}J(\theta | X) =\mathbb{E}_{\pi \sim p_{\theta}(.|X)}[(L(\pi|X)-b(X))\nabla_{\theta}\ln p_{\theta}(\pi|X)]
\end{dmath}

\noindent where $X$ is the CVRP instance\revised{,} and $\pi$ is the solution proposed by the policy. Also, $L(\pi|X)$ is the total cost of the solution $\pi$, that is, the sum of distances. In addition, $b(X)$ is the baseline function for estimating the expected cost of instance~$X$.

This gradient resembles the standard REINFORCE with baseline setup. However, it differs slighlty. In this equation, the weight of the gradient of the log probability is constantly defined, for all the states, as the returns subtracted by the baseline computed for the CVRP instance at hand.

\section{Perturbative Algorithm}
\label{section:perturbative_algorithm}

The perturbative algorithm leveraged in this work is the LSH~\cite{gao2020learn}. The canonical approach tackles both the basic CVRP and the CVRP with time windows by training an algorithm that uses attention mechanisms to learn a local-search heuristic that finds a locally optimal route. This heuristic consists of two operators: \texttt{destroy} and \texttt{repair}. The \texttt{destroy} operator removes some of the nodes from the current solution. In contrast, following the minimum cost principle, the \texttt{repair} operator re-inserts them in different locations of the solution sequence. This approach uses an actor-critic framework for training~\cite{actorcritic}. Such training consists of learning the removal pattern for the \texttt{destroy} operator and the reinsertion order for the \texttt{repair} operator.

The overall model is composed of a modified version of a GAT encoder~\cite{gat2018}, a Gated Recurrent Unit decoder~\cite{cho-etal-2014-learning}, and an actor-critic-based \revised{Reinforcement Learning} algorithm. The modified \revised{GAT} integrates node and edge embeddings for the encoder, which carry information such as the accumulated cost per trip. 

\subsection{Model Encoder}

\revised{The encoder gets the information of each instance through the embeddings. The information from the instances includes information about the nodes and the connections between them. Regarding the information about the nodes, we keep static and dynamic information. The static information concerns the node's demand. Conversely, the dynamic information of a node changes based on the tour that contains it. For each node, we store:

\begin{itemize}
    \item The total demand. The sum of the nodes' demands in the tour that contains the node under analysis.
    \item The accumulated demand. Like the total demand, it also represents the sum of the nodes' demands in the tour. However, it only considers the nodes scheduled for a visit before the node under analysis.
    \item The accumulated distance. The distance traveled up to the node under analysis.
\end{itemize}

Besides, we store information about the connections between each pair of nodes. This information includes their distance and whether or no that connection is part of the solution. These embeddings are projected using a linear transformation and inputted into the EGATE. The activation function of this model is a LeakyReLU, which acts on the concatenation of the two embeddings (the nodes' encodings and the connections' encodings).}

Additionally, the encoder incorporates a masking procedure in which edge embeddings that violate constraints are masked. This model makes up one of the EGATE layers. However, several can be stacked one after the other, where the output of the last one goes into a mean-pooling layer that gives the final output of the encoder.

\subsection{Model Decoder}

The decoder learns the \texttt{destroy} and \texttt{repair} operators, \revised{which define} the nodes to be removed from the solution and their order of reinsertion. It consists of a \revised{Gated Recurrent Unit}, and it is based on the Pointer Network~\cite{NIPS2015_29921001}. The input of each \revised{unit} is the embedding of the node chosen in the previous step, and \revised{its} output foregoes an attention mechanism and a SoftMax layer to produce the next candidate node.

\subsection{Combining Encoder and Decoder}

We use Simulated Annealing~(SA) to combine the encoder and decoder. At every timestep $t$, we compute the embeddings for the current solution. Further such embeddings serve as input to the decoder, which outputs a sequence of nodes. The method generates a new solution by removing all the nodes obtained from the decoder from the current solution and re-inserting them into the solution sequence in the order in which they are presented. Such a re-insertion procedure follows the minimum cost principle. \revised{It inserts e}ach node in the solution sequence at the location where it generates the minimum cost increase. Concretely, at each step, the current solution is updated according to Eq.~\ref{eqn:2}.

\begin{equation}
\label{eqn:2}
    \pi_{t} =
    {\begin{cases}
        \pi_{t},      & \text{if } cost(\pi_{t}) < cost(\pi_{t-1}) - (t_0 \times \alpha^{t}) \ln(\mu_{t}) \\
        \pi_{t-1},              & \text{otherwise}
    \end{cases}}
\end{equation}

\noindent where $\pi$ is the solution, $t$ is the current timestep of the perturbation process, $t_{0}$ is the initial temperature, and $\mu_{t}$ is a continuous uniform random variable between 0 and 1 drawn at every timestep. 

However, in practice, we do not use the $\alpha$ constant by itself. Instead, we compute it as a function of the number of perturbation steps needed to reach a temperature of 1, that is, the product of $(t_{0} \times \alpha^{t})$ becomes 1. Notably, the perturbation steps needed to reach a temperature of 1 as $steps\_t1$ to obtain $\alpha$ is defined by Eq.~\ref{eqn:3}.

\begin{equation}
\label{eqn:3}
    \alpha = (1 / t_0)^{(1 / steps\_t1)}
\end{equation}

\subsection{The Value Network}

The value network used in the actor-critic algorithm follows a simple structure consisting of a two-layered Feed Forward \revised{Neural Network}. The first one is a dense layer with a ReLU activation function and the second one is a linear layer. Its state is defined by the output of the encoder at step $t$ ($Enc^{(t)}$), and the state value is estimated as $v(Enc^{(t)}, \phi)$, where $\phi$ are the parameters of the value network.

\subsection {Training Algorithm}

As described before, the actor-critic algorithm is responsible for the training process. First, we train the critic and then the actor through Proximal Policy Optimization~\cite{DBLP:journals/corr/SchulmanWDRK17}. Notably, the decrease in the overall distance traveled is used as a reward in each CVRP state.

First, in the training process, the advantage is computed as the Temporal-Difference~(TD) error following Eq.~\ref{eqn:4}.

\begin{equation}
\label{eqn:4}
    \delta_{TD}^{(t)} = r^{(t)} + \gamma \times v(Enc^{(t)}, \phi) - v(Enc^{(t-1)}, \phi)
\end{equation}

\noindent where $\gamma$ is the discount factor and $r^{(t)}$ is the reward at timestep $t$. Next, the critic network is trained using Eq.~\ref{eqn:5}.

\begin{equation}
\label{eqn:5}
    \phi = \phi + \eta \times \delta_{TD}^{(t)} \times \nabla_{\phi}v(Enc^{(t)}, \phi)
\end{equation}

\noindent where $\eta$ is the learning rate. The actor is trained through the clipped surrogate objective that optimizes the objective $L^{Clip}(\theta)$ described in Eq.~\ref{eqn:6}.

\begin{dmath}
\label{eqn:6}
    L^{Clip}(\theta) = \mathbb{E}\{min[R_t(\theta) \delta_{TD}^{(t)}, clip(R_t(\theta), 1 - \epsilon,1 + \epsilon) \delta_{TD}^{(t)}]\}
\end{dmath}

\noindent where $R_t(\theta)$ is the ratio of the new policy over the old one. We fixed the value of $\epsilon$ to 0.2 for this work for empirical reasons.

\section{Algorithm Proposal and Improvements}
\label{section:proposal}

In this paper, we propose an algorithm that combines the ones described in the last two sections: the AM-D and the LSH. We demonstrate the effectiveness of the proposed method by \revised{providing outstanding results in common dataset distributions}, as seen in Table~\ref{table:main_results}. Additionally, we present and use \revised{our} optimization for the training algorithm of the AM-D. Such an optimization reduces \revised{its} memory complexity by a factor of the number of nodes in the problem. Furthermore, we also make the code of the CDCP publicly available\footnote{https://github.com/Roberto09/Combined-Deep-Constuctor-and-Perturbator}.

\subsection{Combination of Constructive and the Perturbative Algorithms}

We propose a model that combines the AM-D and LSH, which produces solutions neither of these would be able to generate in isolation. Our inspiration comes from techniques that show success in popular non \revised{Deep Learning} algorithms such as LKH. \revised{In his work,} Helsgaun~\cite{HELSGAUN2000106} describes how the use of intelligently constructed initial solutions can improve, in time or cost, solutions generated through perturbative algorithms. An example of this is that of the Lin-Kernighan algorithm for the \revised{Traveling Salesman Problem} from the Operations Research field. We adapt this idea to the \revised{Machine Learning} domain with our presented method.

To combine the models, we train the AM-D~(constructive model) on the data distribution proposed by Nazari et al.~\cite{nazari2018reinforcement} for CVRP20~(containing 20 nodes and a vehicle capacity of 30), CVRP50~(containing 50 nodes and a vehicle capacity of 40) and CVRP100~(containing 100 nodes and a vehicle capacity of 50). We do this until we reach a cost plateau for the CVRPs. Sequentially, we sample CVRP instances using the same data distribution, and generate solution samples with the trained AM-D. \revised{To obtain these initial solutions, we follow a sampling procedure in which, for every individual CVRP instance, we sample 100 solutions using the model and pick the best one.} Further, we train the LSH~(perturbative model) leveraging the obtained solutions instead of the naively generated ones. \revised{The result is} a model that, on evaluation, intelligently builds a solution and iteratively perturbs it as much as desired or time permits. In the evaluation stage, we apply 1000 of these perturbation operations to the solutions, which is ten times more than what is used for training. Hence, this model can produce solutions at least as good as those proposed by the AM-D and improvements become a function of the time spent in the perturbation phase. The experiments show a significant cost improvement in the solutions compared to the constructive and perturbative models. For clarity, we formally describe the proposed training process for the model in Algorithm~\ref{alg:1}.

\begin{algorithm*}[ht!] \caption{Training Combination of Constructive and Perturbative Models}
\label{alg:1}
\begin{algorithmic}[1]
\Require
\Statex $\beta$, the already trained constructive model
\Statex $\theta$, the un-trained perturbative model
\Statex $bsize$, the batch size used at each training step
\Statex $tsteps$, the train steps; this is similar to the number of epochs
\Statex $rsteps$, number of steps to randomly perturb a given initial solution
\Statex $nnodes$, number of nodes the model perturbs at each step
\Statex $\alpha$, term used in the temperature reduction function for SA
\Statex $\alpha_{rand}$, term used in the temperature reduction function for SA at random initialization
\Statex $perturbations$, number of times the perturbation procedure is applied to a solution sequence
\For{$t \gets 1$ to $tsteps$}
    \State $envs \gets \{\}$
    \For{$b \gets 1$ to $bsize$} \Comment{Obtain solutions using the constructive model}
        \State $vrps_b \gets sample\_vrp\_instance()$
        \State $constructive\_solutions_b \gets \beta(vrps_b)$
        \State $batch_b \gets (vrps_k, constructive\_solutions_b)$
        \State $envs_b \gets create\_environment(batch_b)$
        \State $envs_b \gets initial\_perturbation(envs_b, \alpha_{rand}, rsteps)$
    \EndFor

    \State $states \gets \{\}$
    \For{$p \gets 1$ to $perturbations$} \Comment{Create buffer with perturbation state samples}
        \State $removed\_nodes \gets \{\{\}, \ldots, \{\}\}$
        \For{$n \gets 1$ to $nnodes$}
            \State $removed\_nodes_{b,n} \gets \theta(envs_b)$, $b=1,\ldots,bsize$
            \State $envs_b \gets remove\_node(envs_b, removed\_nodes_{b, n})$, $b=1,\ldots,bsize$
        \EndFor
        \State $envs_b \gets reinsert\_min\_cost\_principle(envs_b, removed\_nodes_b, \alpha)$, $b=1,\ldots,bsize$
        \State $states \gets states \cup (envs_b, removed\_nodes_b)$, $b=1,\ldots,bsize$
    \EndFor
    
    \ForEach{$state\_batch \in split(states, bsize)$} \Comment{Train perturbative model using state buffer}
        \State $envs_b, removed\_nodes_b \gets state\_batch$, $b=1,\ldots,bsize$
        \State $log\_likelihoods_b \gets evaluate(\theta, envs_b, removed\_nodes_b)$, $b=1,\ldots,bsize$
        \State $backpropagate(\theta, log\_likelihoods, state\_batch)$
    \EndFor
\EndFor
\end{algorithmic}
\end{algorithm*}

As seen in Algorithm~\ref{alg:1}, for every $tstep$, we sample a new CVRP instance and solve it with the AM-D. However, in practice, we pre-compute a finite number of them and their solutions. We refer to the amount of such pre-computed instances as $ninstances$. Notably, this algorithm applies an initial random perturbation to the generated solutions in line 8. This initial random perturbation is part of the LSH implementation\revised{, which we formally describe} in Algorithm~\ref{alg:2}.

\begin{algorithm*}[ht!] \caption{Initial Perturbation of a Batch of Constructive Solutions}
\label{alg:2}
\begin{algorithmic}[1]
\Require
\Statex $envs$, environments of batch with CVRP solutions obtained from the constructive model
\Statex $rsteps$, number of steps to randomly perturb the initial solution
\Statex $\alpha_{rand}$, term used in the temperature reduction function of SA at random initialization
\Statex $N$, total number of nodes in the CVRP
\For{$i \gets 1$ to $rsteps$}
\State $removed\_nodes_k \gets uni\_rnd\_smpl\_no\_repl(n=10, \{1,\ldots,N\})$, $k=1,\ldots,|envs|$
\State $envs_k \gets remove\_node(envs_k, removed\_nodes_k)$, $k=1,\ldots,|envs|$
\State $envs_k \gets reinsert\_min\_cost\_principle(envs_k, removed\_nodes_k, \alpha_{rand})$, $k=1,\ldots,|envs|$
\EndFor
\end{algorithmic}
\end{algorithm*}

\subsection{Gradient Computation Improvement for AM-D}

In the first iteration of the constructive model proposed by Kool et al., the authors mentioned they faced memory constraints~\cite{kool2019attention}. Consequently, the authors had to reduce the batch size from 512 to 256 instances in experiments for CVRPs with 100 nodes. They conducted experiments using two GTX-1080 TI, which together have a GPU memory of 22 GB. Similarly, on a second iteration of the model proposed by Peng et al., the authors reduced the batch size of CVRPs with 100 nodes from 128 to 108 instances~\cite{peng2020deep}. Although they omitted the reasons, we think it is related to memory issues. Likewise, we ran into such a memory wall in our experiments for this work. This issue was prominent when running training on one of our available servers with 6 GB of GPU memory.

\begin{algorithm*}[ht!]
\caption{Original Constructive Model Backpropagation on Batch}
\label{alg:3}
\begin{algorithmic}[1]
\Require
\Statex $\theta$, the constructive model weights
\Statex $p_{\theta}$, the constructive model
\Statex $batches$, a list of batches with unsolved CVRPs
\Statex $\eta$, the learning rate
\ForEach {$batch \in batches $}
\State $log\_likelihoods \gets \{0, \ldots, 0\} $
\State $costs \gets \{0, 0, \ldots, 0\} $
\State $baseline\_costs \gets eval\_baseline(batch) $
\State $env \gets create\_environments(batch)$
\While{$\neg all\_finished(env)$}
\State $ll, selected\_node, cost \gets p_{\theta}(env)$
\State $log\_likelihoods_i \gets log\_likelihoods_i + ll_i$, $i=1,\ldots,|batch|$
\State $costs_i \gets costs_i + cost_i$, $i=1,\ldots,|batch|$
\State $step\_env(env, selected\_node)$
\EndWhile
\State $\nabla_{\theta}J(\theta) \gets \frac{1}{|batch|} \sum_{i=1}^{|batch|} (costs_i - baseline\_costs_i) \times \nabla_{\theta}\ log\_likelihoods_i$
\State $\theta_j \gets \theta_j - \eta \times \nabla_{\theta}J(\theta)_j$, $j=1,\ldots,|\theta|$
\EndFor
\end{algorithmic}
\end{algorithm*}

To solve this problem, we made an observation in Eq.~\ref{eqn:1}, which is implemented initially as Algorithm~\ref{alg:3}. We noticed that computing the required gradient is extremely expensive memory-wise due to the $\nabla_{\theta}\ln p_{\theta}(\pi|X)$ calculation. For this, we have to compute and accumulate the computational graphs, that calculate the log probabilities at each step, in memory. Hence, the number of computational graphs we must store in memory is a function of the number of nodes in the CVRP. Such a task consumes plenty of GPU memory given the way PyTorch 1.10.0 \cite{Paszke2017AutomaticDI} works. In this library, the constructed computational graph used for backpropagation will keep multiple tensors in GPU memory. These tensors are all the inputs and intermediate results obtained while computing the suggested nodes for every step\revised{, which increases memory consumption} as the size of the CVRPs \revised{grow}.

\subsubsection{Memory consumption and runtime analysis}

Analyzing the AM-D, the total memory allocated when we compute the loss gradient is in the order \revised{given by} Eq.~\ref{eqn:10}.

\begin{equation}
\label{eqn:10}
    n \times (m_{1} \times n^{2} + m_{2} \times n + m_{3})
\end{equation}

\noindent where $n$ is the number of nodes in the CVRP and $m_{1}$, $m_{2}$ and $m_{3}$ are constants related to the embedding sizes, number of weights used in layers, number of layers, etc.

Similarly, at runtime, the total number of operations to compute the gradient of the loss function has the same polynomial form as the memory allocation. This is in the order of Eq.~\ref{eqn:11}.

\begin{equation}
\label{eqn:11}
    n \times (r_{1} \times n^{2} + r_{2} \times n + r_{3})
\end{equation}

\noindent where $n$ is the number of nodes in the CVRP and $r_{1}$, $r_{2}$ and $r_{3}$ are constants similar to those of Eq.~\ref{eqn:10}.

For both the memory allocation and runtime approximation, we included the quadratic term $n^{2}$ as part of Eqs.~\ref{eqn:10} and~\ref{eqn:11}. We add this term for cases \revised{similar to} the encoder\revised{'s}, where we compute the MHA of a given set of nodes. For this, in every head, we make use of an attention mechanism that involves the product of a query matrix $q \in \mathbb{R}^{n,c}$ and a key matrix $k \in \mathbb{R}^{c,n}$. This product results in a matrix representing the attention weights. For this, we must allocate an n by n matrix and make a total of $c \times n^{2}$ operations, where $c$ is a constant representing the embedding sizes. We also multiply the $n^{2}$ term by $m_{1}$ and $r_{1}$ in the memory and runtime equations, respectively. We include these multiplicative constants to account for things like multiple instances in the batch, various heads in an MHA layer, several MHA stacked layers\revised{,} or other cases where we must store or compute an $n^{2}$ sized matrix.

Similar to the quadratic term case, we included the linear term $n$ as part of Eqs.~\ref{eqn:10} and~\ref{eqn:11}. \revised{We include t}his term for cases such as that of the decoder\revised{,} where we compute the context vector. For this, in every head of the MHA, we apply an attention mechanism that involves the product of a query vector $q \in \mathbb{R}^{c}$ and a key matrix $k \in \mathbb{R}^{c,n}$. This \revised{process} results in a vector representing the attention weights. For this, we must allocate a 1-by-$n$ vector and make a total of $c \times n$ operations, where $c$ is a constant. We also multiply the $n$ term by $m_{2}$ and $r_{2}$ in the memory and runtime equations, respectively. We include these multiplicative constants to account for cases similar to those of $m_{1}$ and $r_{2}$ in the $n^{2}$ case.

For completeness, we also include the $m_{3}$ and $r_{3}$ constants as part of Eqs.~\ref{eqn:10} and~\ref{eqn:11}. This accounts for various operations resulting in constant-sized allocations or computations. An example is that of the decoder, where we compute the context vector. For this, we must make a linear projection in every head of the MHA. In the linear projection, we multiply the resulting vector of the attention mechanism $a \in \mathbb{R}^{c}$ by a weight matrix $w \in \mathbb{R}^{c,c}$. Consequently, we must allocate a $c^{2}$ matrix and make a total of $c^{3}$ operations, where c is a constant.

Finally, for both the memory allocation and runtime approximations, we multiply everything inside the parentheses by the term $n$, as seen in Eqs.~\ref{eqn:10} and~\ref{eqn:11}. We include this term since we run the model approximately $n$ times to produce all the CVRP solutions\revised{,} hence the loss gradient. For each of those steps, we must keep in memory all the tensors we allocate since they are needed by the computational graph to run backpropagation at each iteration.

Notably, the encoder (responsible for the $n^{2}$ term inside the parentheses) only runs when an instance’s partial solution is at the depot. Whether or not the times the encoder runs is a function of the number of steps we make is arguable. If it is desired not to consider it, then the above analysis is slightly modified by removing $n^{2}$ from inside the parentheses in both the memory allocation and runtime approximations.

In big $\mathcal{O}$ notation, we express both the memory and runtime complexity as $\mathcal{O}(n^{3})$.

\subsubsection{Memory optimization}

We observed that, by rearranging the gradient equation to a different form, we can better instruct PyTorch to compute the gradients in a way that reduces the memory allocation to only $m_{1} \times n^{2} + m_{2} \times n + m_{3}$. We achieve this by doing the forward pass twice and backpropagating at every solution step resulting in $3n \times (r_{1} \times n^{2} + r_{2} \times n + r_{3})$ operations.

From the big $\mathcal{O}$ perspective, this is an excellent tradeoff as we reduce the memory complexity to $\mathcal{O}(n^{2})$ without impacting the runtime complexity. To accomplish this, we start from the original gradient approximation equation for the policy $p_{\theta}$ described in Eq.~\ref{eqn:14}.

\begin{equation}
\label{eqn:14}
    \nabla_{\theta}J(\theta) \approx \frac{1}{B} \sum_{i=1}^B [\Delta{L_i} \nabla\ln p_{\theta}(\pi_i^s|X_i)]
\end{equation}

\noindent where

\begin{equation}
\label{eqn:14b}
    \Delta{L_i} = L(\pi_i^s|X_i) - L(\pi_i^g|X_i)
\end{equation}

\noindent where $\pi_{i}^{s}$ and $\pi_{i}^{g}$ are the solutions of the CVRP instance $X_{i}$. In this computation, $\pi_{i}^{s}$ is obtained by performing sampling and $\pi_{i}^{g}$ by greedy search using the policy. $L(\pi|X_{i})$ is the function that computes the total cost, that is, the sum of distances of a given solution $\pi$. Furthermore, $B$ is the batch size of $X$.

We expand the definition of $\nabla\ln p_{\theta}(\pi_i^s|X_{i})$ to get Eq.~\ref{eqn:15}.

\begin{dmath}
\label{eqn:15}
    \nabla_{\theta}J(\theta) \approx \frac{1}{B} \sum_{i=1}^B [\Delta{L_i} \times \nabla_{\theta} \sum_j^{|\pi_i^s|}\ln p_{\theta}(\pi_{i,j}^s|X_i)]
\end{dmath}

By swapping the sums, we obtain Eq.~\ref{eqn:16}.

\begin{dmath}
\label{eqn:16}
    \nabla_{\theta}J(\theta) \approx \frac{1}{B} \sum_j^{\max_k |\pi_k^s|} \nabla_{\theta}  \sum_{i=1}^B
    {\begin{cases}
        [\Delta{L_i} \times \ln P_{i,j}],& \text{if } j \leq \max_k |\pi_k^s|\\
        0,              & \text{otherwise}
    \end{cases}}
\end{dmath}

\noindent where
\begin{equation}
\label{eqn:16b}
    {P_{i,j}} = p_{\theta}(\pi_{i,j}^s|X_i)
\end{equation}

Finally, we reach the desired form in Eq.~\ref{eqn:17}.

\begin{dmath}
\label{eqn:17}
    \nabla_{\theta}J(\theta) \approx \sum_j^{\max_k |\pi_k^s|} \frac{1}{B} \sum_{i=1}^B
    {\begin{cases}
        [\Delta{L_i} \times \nabla_{\theta} \ln P_{i,j}],& \text{if } j \leq \max_k |\pi_k^s|\\
        0,              & \text{otherwise}
    \end{cases}}
\end{dmath}

Translating this gradient computation into an algorithm allows us to use PyTorch's gradient accumulation functionality. This translation is advantageous, as opposed to the original, since we can backpropagate every time we compute a new node to be added to the solution. With this change, gradients will accumulate for us across steps (outer sum in Eq.~\ref{eqn:17}) every time we do a backward pass. For this reason, we do not need to store the intermediate computational graphs every time we compute a new step of the solution.

Nevertheless, we need the cost difference between the obtained solution and the baseline \revised{$\Delta{L_i}$} before we can backpropagate at a given step. This term is required since \revised{it} is multiplying the gradient of the log probability \revised{that} the model computed for the selected node in Eq.~\ref{eqn:17}. To address this, we first run the model to produce the selected nodes with the gradient computation turned off. Further, we compute the proposed solution cost and the baseline cost. Then we use the selected nodes and costs to compute the gradient by doing a second forward pass with the gradient computations turned on. This means we  multiply the previous runtime by three since we perform two forward passes and one backpropagation at every step.

The implementation of the memory-efficient procedure is described in Algorithm~\ref{alg:4}.

\begin{algorithm*}[ht!]
\caption{Memory-Efficient Constructive Model Backpropagation on Batch}
\label{alg:4}
\begin{algorithmic}[1]
\Require
\Statex $\theta$, the constructive model weights
\Statex $p_{\theta}$, the constructive model
\Statex $batches$, a list of batches with unsolved CVRPs
\Statex $\eta$, the learning rate
\ForEach {$batch \in batches $}
\State $costs \gets \{0, \ldots, 0\} $
\State $env \gets create\_environments(batch)$
\While{$\neg all\_finished(env)$} \Comment{First pass to compute costs}
\State $ll, selected\_node, cost \gets p_{\theta}(env)$
\State $costs_i \gets costs_i + cost_i$, $i=1,\ldots,|batch|$
\State $env \gets step\_env(env, selected\_node)$
\EndWhile
\State $env \gets create\_environments(batch)$
\State $baseline\_costs \gets eval\_baseline(batch) $
\State $\nabla_{\theta}J(\theta) \gets \{0, \ldots, 0\} $
\While{$\neg all\_finished(env)$} \Comment{Second pass to compute gradients}
\State $ll, selected\_node, cost \gets p_{\theta}(env)$
\State $env \gets step\_env(env, selected\_node)$
\State $grads \gets \frac{1}{|batch|} \sum_{i=1}^{|batch|} (costs_i - baseline\_costs_i) \times \nabla_{\theta} \ ll_i$
\State $\nabla_{\theta}J(\theta)_j \gets \nabla_{\theta}J(\theta)_j + grads_j$, $j=1,\ldots,|\theta|$
\EndWhile
\State $\theta_j \gets \theta_j - \eta \times \nabla_{\theta}J(\theta)_j$, $j=1,\ldots,|\theta|$
\EndFor
\end{algorithmic}
\end{algorithm*}

\section{Experiments}
\label{section:experiments}

We conduct experiments to assess the performance of the CDCP and the memory consumption improvement of the AM-D. Concretely, we experiment with various hyperparameters of the CDCP, using CVRPs with 50 nodes, to find an optimal configuration. Further, we use \revised{that} configuration in CVRPs of 20, 50, and 100 nodes to test its overall performance. Additionally, we test the memory consumption improvement of our optimized AM-D training algorithm \revised{on} various CVRP configurations. We also illustrate the runtime tradeoff when using this algorithm. Furthermore, we provide a discussion of all the experiments we carried out.

\subsection{Combination of Constructive and Perturbative Algorithms}

\begin{figure*}
    \centering
    \includegraphics[width=\linewidth]{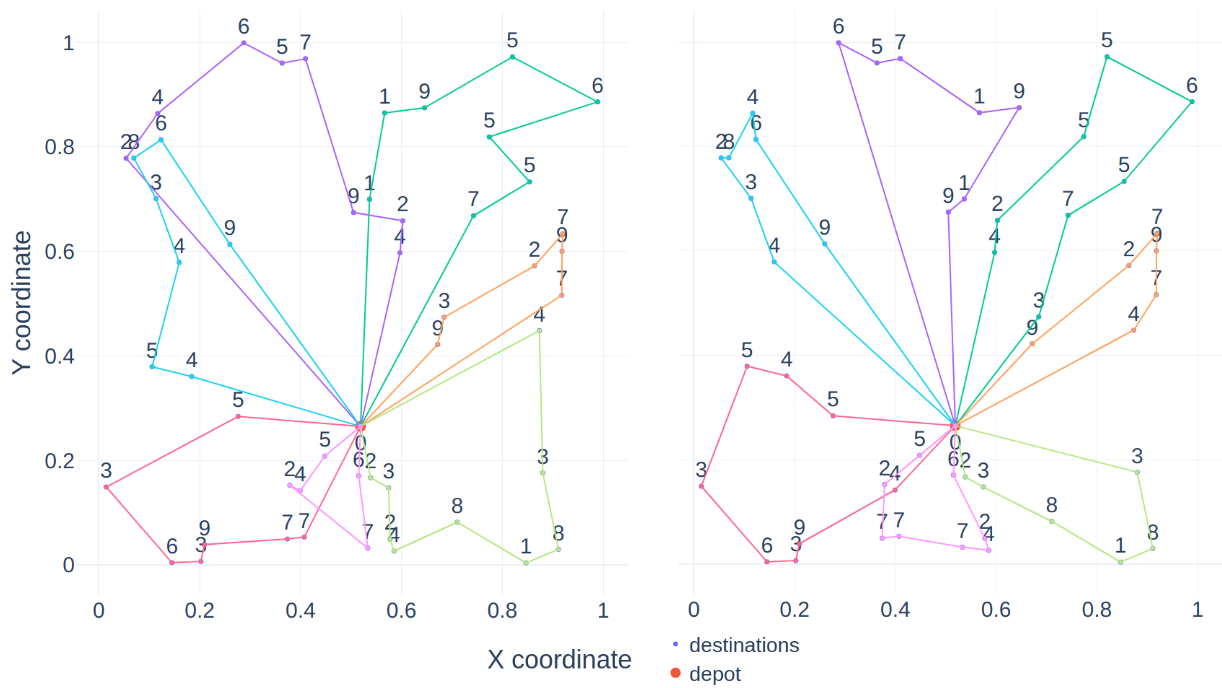}
    \caption{Solution for a CVRP50 using the AM-D with sampling (on the left) and solution using the CDCP (on the right). The AM-D obtained a solution with a cost of 10.17 while the CDCP obtained a cost of 9.46. Each colored line represents a tour traveled by the vehicle, and the points represent the locations and the depot. In addition, the number on top of each location is its demand.}
    \label{fig:cdcp_sample}
\end{figure*}

\begin{figure*}[htb!]
    \centering
    \includegraphics[width=\linewidth]{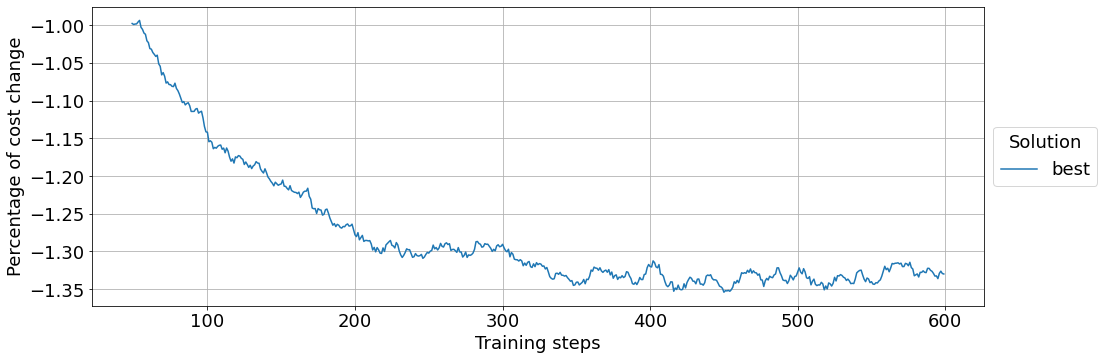}
    \caption{\revised{Percentage of cost change of the solutions obtained after the perturbation process on CVRP20 instances. Solutions are gathered during the training process and their cost change is obtained by comparing the initial cost and the cost of the best solution obtained by CDCP. Here, we define the costs as the total distance traveled by the vehicle on a given instance. The percentage of cost change is plotted as the rolling mean over the last 50 training steps.}}
    \label{fig:5}
\end{figure*}

\begin{figure*}[htb!]
    \centering
    \includegraphics[width=\linewidth]{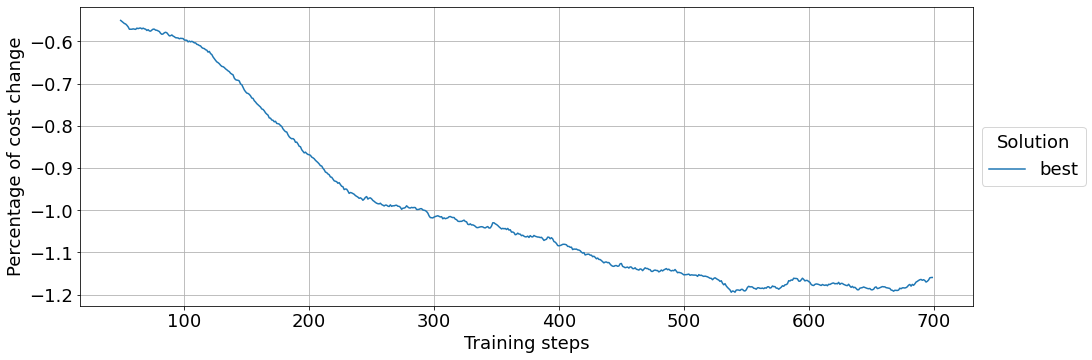}
    \caption{Percentage of cost change of the solutions obtained after the perturbation process on CVRP50 instances. Solutions are gathered during the training process of the CDCP. The percentage of cost change is plotted as the rolling mean over the last 50 training steps.}
    \label{fig:6}
\end{figure*}

\begin{figure*}[htb!]
    \centering
    \includegraphics[width=\linewidth]{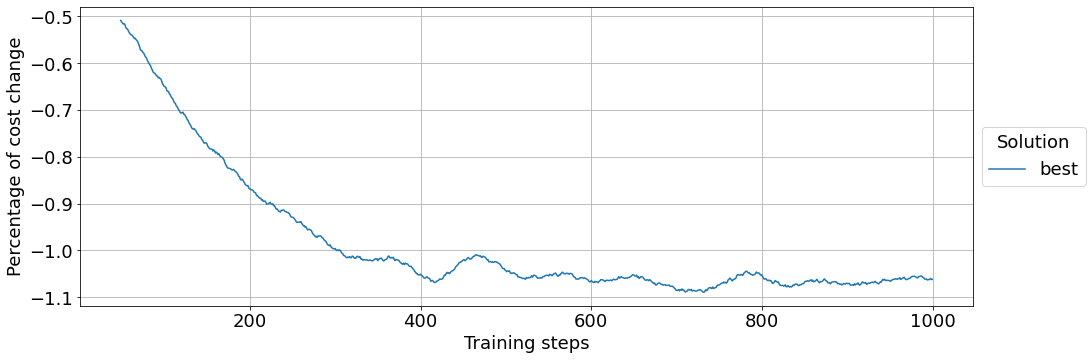}
    \caption{Percentage of cost change of the solutions obtained after the perturbation process on CVRP100 instances. Solutions are gathered during the training process of the CDCP. The percentage of cost change is plotted as the rolling mean over the last 50 training steps.}
    \label{fig:7}
\end{figure*}

We report the \revised{cost and time} results obtained by the CDCP in Table~\ref{table:main_results}. As it is common in the literature, we obtain our results by sampling 10K CVRP instances from the data distribution proposed by Nazari et al.~\cite{nazari2018reinforcement} and running CDCP over them. For all the CDCP experiments, first we trained the AM-D with the configurations \revised{shown} in Table~\ref{table:AMDHyperParameters}. Further, we sampled CVRPs and generated their solutions using the constructive algorithm. Finally, we trained the LSH using the generated instances and solutions with the configurations listed in Table~\ref{table:LSHHyperParameters} and following Algorithm~\ref{alg:1}. We present the training results of these models in Figs.~\ref{fig:5},~\ref{fig:6} and~\ref{fig:7}. Additionally, we provide an example of a solution for a CVRP50 instance in Fig.~\ref{fig:cdcp_sample}.

\begin{table*}[ht!]
\caption{Results on CVRP. \revised{Cost} is the average length of the test \revised{solutions}. Gap is the optimality gap computed as the percentage cost change over the state-of-the-art results. \revised{Time is the total time required by each model to solve all the test instances}.}
\label{table:main_results}
\centering
\begin{tabular*}{1\textwidth}{@{}@{\extracolsep{\fill}}*{10}{c}@{}}
\toprule
\multirow{2}{*}{\textbf{Method}} & \multicolumn{3}{c}{\begin{tabular}[c]{@{}c@{}}\textbf{CVRP20}\\ Capacity 30\end{tabular}} & \multicolumn{3}{c}{\begin{tabular}[c]{@{}c@{}}\textbf{CVRP50}\\ Capacity 40\end{tabular}} & \multicolumn{3}{c}{\begin{tabular}[c]{@{}c@{}}\textbf{CVRP100}\\ Capacity 50\end{tabular}} \\ \cmidrule(l){2-10} 
 & Cost & Gap (\%) & Time & Cost & Gap (\%) & Time & Cost & Gap (\%) & Time \\ \midrule
\revised{HGS} \cite{DBLP:journals/corr/abs-2012-10384} & - & - & - & - & - & - & 15.56 & 0.00 & 6h 11m \\
Gurobi LLC 2018 & 6.10 & 0.00 & - & -& - & - & - & - & - \\
LKH3 \cite{HELSGAUN2000106} & 6.14 & 0.66 & 2h & 10.38 & 0.29 & 7h & 15.65 & 0.58 & 13h \\
\begin{tabular}[c]{@{}c@{}}\revised{Reinforcement Learning}\\ for the VRP \cite{nazari2018reinforcement}\end{tabular} & 6.40 & 4.92 & - & 11.15 & 7.73 & - & 16.96 & 9.00 & - \\
\begin{tabular}[c]{@{}c@{}}Learning Based\\ Iterative Method~\cite{Lu2020A}\\ (Computed only\\ for 2000 instances)\end{tabular} & 6.12 & 0.33 & - & 10.35 & 0.00 & - & 15.57 & 0.06 & - \\
\begin{tabular}[c]{@{}c@{}}Learning to Perform\\ Local Rewriting~\cite{chen2019learning}\end{tabular} & 6.16 & 0.98 & 22m & 10.51 & 1.55 & 18m & 16.10 & 3.47 & 1h \\
AM (sampling)~\cite{kool2019attention} & 6.25 & 2.46 & 6m & 10.62 & 2.61 & 28m & 16.23 & 4.31 & 2h \\
AM-D (greedy)~\cite{peng2020deep} & 6.28 & 2.95 & 3s & 10.78 & 4.15 & 25s & 16.40 & 5.40 & 2m 39s \\
Deep Policy Dynamic\\ Programming (1M)~\cite{kool2021deep} & - & - & - & - & - & - & 15.63 & 0.45 & 48h 37m \\
\textbf{CDCP} & 6.13 & 0.49 & 4h 53m & 10.47 & 1.16 & 10h 12m & 15.85 & 1.86 & 19h 22m \\ \bottomrule
\end{tabular*}
\end{table*}

\begin{table*}[htb!]
\caption{Hyper-parameters used for training the AM-D.}
\label{table:AMDHyperParameters}
\centering
\begin{tabular*}{0.8\textwidth}{@{}@{\extracolsep{\fill}}*{5}{c}@{}}
    \toprule
    & \textbf{Hyper-parameter} & \textbf{CVRP20} & \textbf{CVRP50} & \textbf{CVRP100} \\ \midrule
    \multirow{6}{*}{\textbf{Training}} & \textit{samples\_per\_epoch} & 512 $\times$ 2500 & 512 $\times$ 2500 & 256 $\times$ 2500 \\
    & \textit{bsize} & 512 & 512 & 256 \\
    & \textit{epochs} & 146 & 65 & 100 \\
    & \textit{$\eta$} & $3 \times 10^{-4}$ & $3 \times 10^{-4}$ & $3 \times 10^{-4}$ \\
    & \textit{mem\_efficient} & False & True & True \\ \midrule
    \textbf{Model} & \textit{embedding\_dim} & 128 & 128 & 128 \\ \bottomrule
\end{tabular*}
\end{table*}

\begin{table*}[htb!]
\caption{Hyper-parameters used for training the LSH.}
\label{table:LSHHyperParameters}
\centering
\begin{tabular*}{0.8\textwidth}{@{}@{\extracolsep{\fill}}*{5}{c}@{}}
\toprule
 & \textbf{Hyper-parameter} & \textbf{CVRP20} & \textbf{CVRP50} & \textbf{CVRP100} \\ \midrule
\textbf{Training} & \textit{ninstances} & 13056 & 13056 & 13056 \\
 & \textit{bsize} & 128 & 128 & 128 \\
 & \textit{tsteps} & 600 & 700 & 1000 \\
 & \textit{buses} & 4 & 4 & 4 \\
 & \textit{rsteps} & 0 & 0 & 0 \\ 
 & \textit{$\eta$} & $3 \times 10^{-4}$ & $3 \times 10^{-4}$ & $3 \times 10^{-4}$ \\ \midrule
\multirow{2}{*}{\textbf{Model}} & \textit{nnodes} & 10 & 10 & 10 \\ 
 & \textit{perturbations} & 100 & 100 & 100 \\ \midrule
\multirow{5}{*}{\textbf{\begin{tabular}[c]{@{}c@{}}Simulated\\ Annealing\end{tabular}}} & \textit{$t_0$} & 100 & 100 & 100 \\
 & \textit{steps\_t1} & 1 & 1 & 1 \\
 & \textit{$\alpha$} & $ (1 / t_{0})^{(1/steps\_t1)} $  & $ (1 / t_{0})^{(1/steps\_t1)} $ & $ (1 / t_{0})^{(1/steps\_t1)} $ \\
 & \textit{$\alpha_{rand}$} & 0 & 0 & 0 \\ \bottomrule
\end{tabular*}
\end{table*}

In our proposal, we change a strong prior of the LSH, which is that it starts the perturbation process from far-from-good solutions. For this reason, we decided to experiment with three main knobs. These are the amount of random perturbation of the initial solutions~(generated by the AM-D), the conservativeness degree of the SA used to sequentially perturb the solutions, and the number of nodes we perturb every time the model is run on a CVRP instance. We performed the following experiments in CVRP50 with the hyper-parameters listed in Table~\ref{table:LSHHyperParameters} (unless stated otherwise in the experiment description).

\subsubsection {Number of Perturbed Nodes At Each Model Step}

The LSH applies the perturbation procedure over a given solution for several steps. We experimented with five values, ranging from 2 to 15, of the number of nodes to remove and re-insert in each perturbation step. We refer to this quantity as the number of perturbed nodes. \revised{L}ines 11 to 19 of Algorithm~\ref{alg:1} \revised{formally describe this perturbation process}.

\begin{figure*}[ht!]
    \centering
    \includegraphics[width=\linewidth]{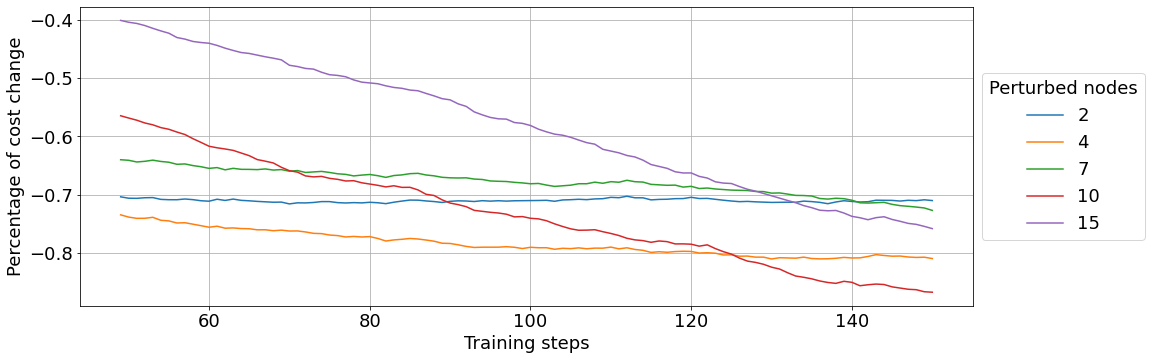}
    \caption{\revised{Percentage of cost change of the solutions obtained after the perturbation process using different amounts of perturbed nodes.} Each line represents a CDCP model with a different number of perturbed nodes, as indicated by its legend. The percentage of cost change is plotted as the rolling mean over the last 50 training steps.}
    \label{fig:perturb_nodes_exp_1}
\end{figure*}

Concretely, the hyper-parameters used for this experiment are the same as those listed in Table~\ref{table:LSHHyperParameters} except for the number of perturbed nodes and the number of steps it takes for the SA to reach temperature 1. For this, we make $nnodes$ take the values of 2, 4, 7, 10, and 15, and $steps\_t1$ always take the value of 1 (based on empirical findings). \revised{This configuration allowed us} to minimize the effect of SA in this experiment \revised{and} test the effect of different numbers of perturbed nodes in isolation. The results of this experiment can be observed in Fig.~\ref{fig:perturb_nodes_exp_1}.

Low values of perturbed nodes do not work as well as higher values of this variable. We observe this when the number of perturbed nodes falls between \revised{2} and \revised{7}. \revised{It may seem obvious that} it might be more challenging for the model to express complex perturbation operations \revised{considering only} a few nodes. \revised{However}, the highest number of perturbed nodes also under-performs, as observed when we perturb\revised{ed} 15 nodes. This result makes us think \revised{that} high values might add unnecessary complexity\revised{,} reflect\revised{ing} lower model quality. A value of 10 \revised{nodes to perturb at each step} (20\% of the \revised{total}) \revised{performed well for this work}.

\subsubsection{Simulated Annealing Conservativeness and Initial Random Perturbation}

The LSH allows using SA during the perturbation process of a solution. It also permits an initial random perturbation phase. The effects of these procedures are closely related \revised{since they affect the} solutions the model \revised{observes} throughout training. For the random initialization case, it is clear from Algorithm~\ref{alg:2} that the solutions the model is trained with will be perturbed concerning those generated by the constructive model. In the case of SA, \revised{solutions are randomly perturbed throughout the training process due to the exponentially decreasing probability of accepting solutions worse than the one at hand in each step}, as seen in Eq.~\ref{eqn:2}.

Due to the \revised{aforementioned} close relationship between the random initialization and SA procedures, we experimented with both simultaneously. We do this by randomly sampling these values from their appropriate distributions. In the case of the number of random initialization steps, we sampled from a uniform discrete distribution between 1 and 20. For the SA case, we sampled the number of steps it takes for the temperature to reach the value of 1 from a distribution defined by ${\lfloor {10^{(\mu)}} \rfloor}$ where $\mu$ is continuously uniformly distributed between 0 and 2. We experimented with 15 samples obtained by following that procedure. Additionally, we manually added four samples, all with 0 random initialization steps, that is, no random initialization, and the number of steps it takes for the SA temperature to reach 1 ranging from 1 to 100.

\begin{figure}[ht!]
    \centering
    \includegraphics[width=\linewidth]{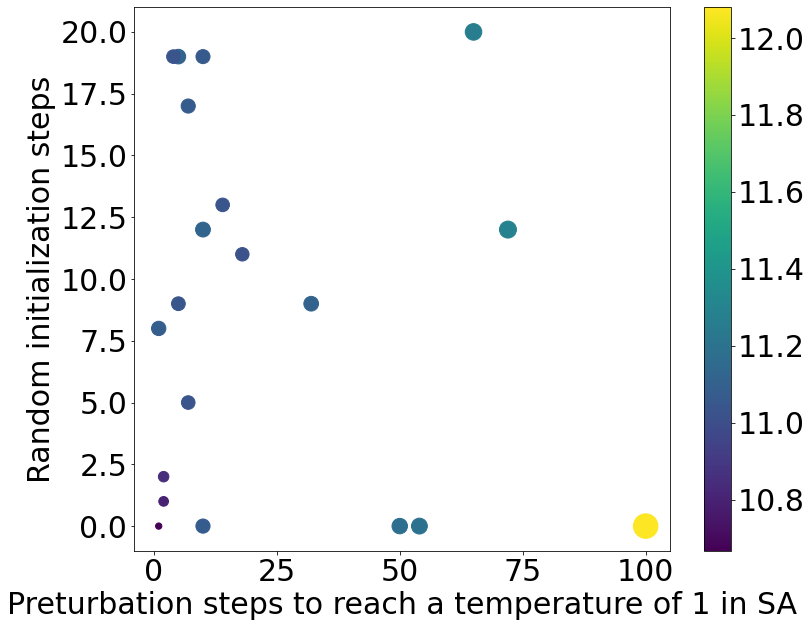}
    \caption{$rsteps$ vs $steps\_t1$ where each point represents a CDCP model with a different number of random initialization steps and steps to reach temperature 1 in SA. The color and size of each point represents the last cost obtained after the perturbation process. The cost is plotted as the mean cost of all the solutions in the last 20 training steps after training the CDCP.}
    \label{fig:sa_cons_and_irp_exp_1}
\end{figure}

\begin{figure*}[ht!]
    \centering
    \includegraphics[width=\linewidth]{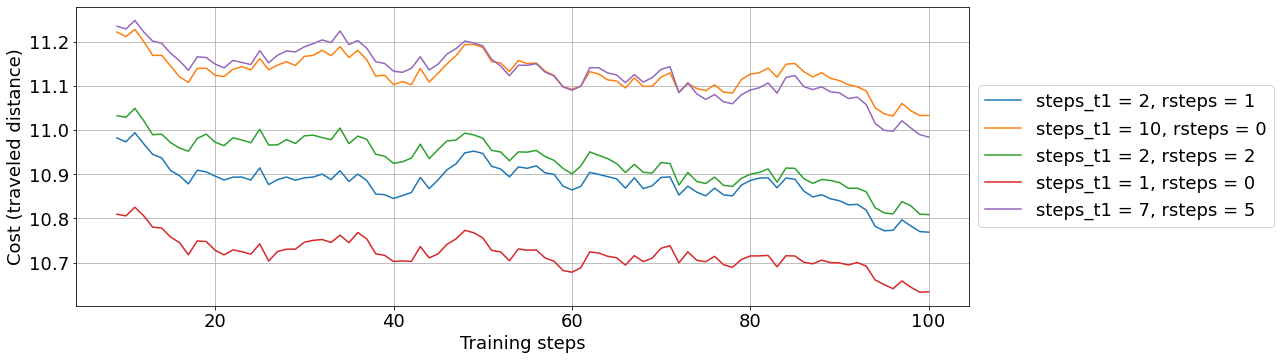}
    \caption{Mean of the last costs obtained after the perturbation process of the models in Fig.~\ref{fig:sa_cons_and_irp_exp_1} with $steps\_t1 \leq 40$ and $rsteps \leq 7.5$. Each line represents a CDCP model with a different number of steps to reach temperature 1 in SA and random initialization steps, as indicated by its legend. Costs are plotted as the rolling mean over the last 50 training steps.}
    \label{fig:sa_cons_and_irp_exp_2}
\end{figure*}

In concrete, the hyper-parameters used for this experiment are the same as those listed in Table~\ref{table:LSHHyperParameters} except for the random initialization steps, the number of steps it takes for the SA to reach temperature 1, and the alpha value used in the SA for random initialization. Particularly, we changed the corresponding hyper-parameters to those listed in Table~\ref{table:2}. The results of this experiment can be observed in Figs.~\ref{fig:sa_cons_and_irp_exp_1} and~\ref{fig:sa_cons_and_irp_exp_2}.

\begin{table*}[ht!]
\caption{SA Conservativeness and Initial Random Perturbation hyper-parameters.}
\label{table:2}
\centering
\begin{tabular*}{0.65\textwidth}{@{}@{\extracolsep{\fill}}cc@{}}
    \toprule
    Hyper-parameter & Values \\
    \midrule
    $rsteps$ & $\{\mu_0, \mu_1, \ldots, \mu_{15}\} \cup \{0,0,0,0\} \mid \mu_i \sim \lfloor \mathcal{U}(0, 20) \rfloor$ \\ 
    $\alpha_{rand}$ & 0.955 \\
    $steps\_t1$ & $ \{\chi_0, \chi_1, \ldots, \chi_{15}\} \cup \{1,10,50,100\} \mid
    \chi_i \sim \lfloor 10^{\mathcal{U}(0, 2)} \rfloor$\\ 
    \bottomrule
\end{tabular*}
\end{table*}

\begin{figure*}[ht!]
    \centering
    \includegraphics[width=\linewidth]{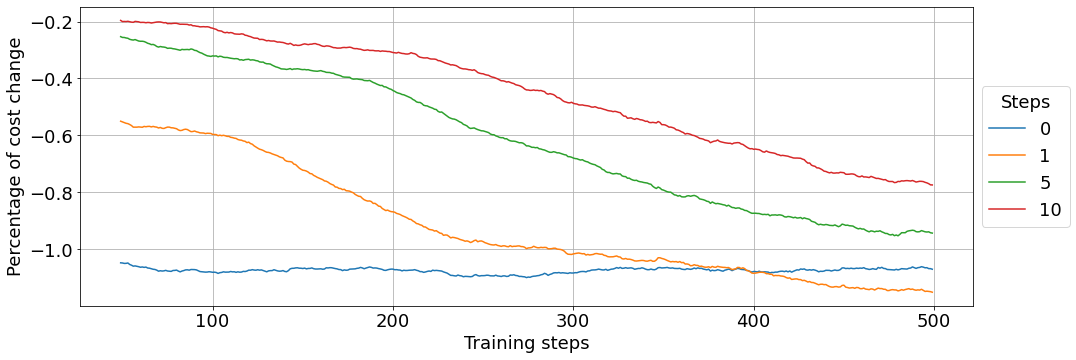}
    \caption{\revised{Percentage of cost change of the solutions obtained after the perturbation process} with no random initialization steps, and steps to reach temperature 1 in SA ranging from 0 to 10. Each line represents a CDCP model with a different number of steps to reach temperature 1 in SA, as indicated by its legend. The percentage of cost change is plotted as the rolling mean over the last \revised{50} training steps.}
    \label{fig:sa_cons_and_irp_exp_3}
\end{figure*}

Additionally, we experimented with four other configurations of the random initialization steps and the number of steps it takes for the SA temperature to reach 1. However, in this case, we ran the experiments for 500 train steps instead of only 100. We did this to test the long-term effect of using SA during training. None of these configurations had an initial random perturbation. However, the number of steps required to reach a temperature of 1 in SA was different for each of them, and remained within the range of 0 and 10. The results of this experiment can be observed in Fig.~\ref{fig:sa_cons_and_irp_exp_3}.

It is clear that low values of both the number of random initialization steps and the number of steps it takes for the SA temperature to reach 1 result in the best perturbation costs after 100 training steps. However, it is unclear from Fig.~\ref{fig:sa_cons_and_irp_exp_1} which values surrounding the ($steps\_t1 = 0$, $rsteps = 0$) point have the best long-term convergence. This occurs since the models in each experiment start learning from solutions with different costs\revised{, as} seen in Fig.~\ref{fig:sa_cons_and_irp_exp_2}\revised{,} where the costs reported at the \revised{first step} of training \revised{of} a sub-sample of the experiments are different. Notably, we observed that a slight perturbation through SA allowed the model to find better solutions when it was trained for longer. In this case, setting the number of steps to reach a temperature of 1 in SA to 1 results in a bigger relative decrease in solution cost than those obtained by other values, as seen in Fig.~\ref{fig:sa_cons_and_irp_exp_3}. Moderately using SA seems beneficial for long-term training as, even when worse solutions are initially expected, it helps in solution exploration throughout training. Additionally, it appears appropriate to not leverage an initial random perturbation.

\subsection{Gradient Computation Improvement for AM-D}

We decided to test the gradient computation improvement with various CVRP settings, including those proposed by Nazari et al.~\cite{nazari2018reinforcement} in an Nvidia Tesla V100 GPU. The CVRPs’ settings used ---by considering tuples (nodes in graph, vehicle capacity)--- are described as follows: (20, 20), (35, 35), (50, 40), (75, 45), and (100, 60).

For this experiment, we trained each CVRP model using embeddings of size 128 for one epoch of 100 batches with 128 instances each. For each CVRP model, we computed the peak CUDA memory consumption using PyTorch statistics. Additionally, we obtained the mean CPU time elapsed across batches using the PyTorch profiler library.

\begin{figure*}[t]
    \centering
    \includegraphics[width=\linewidth]{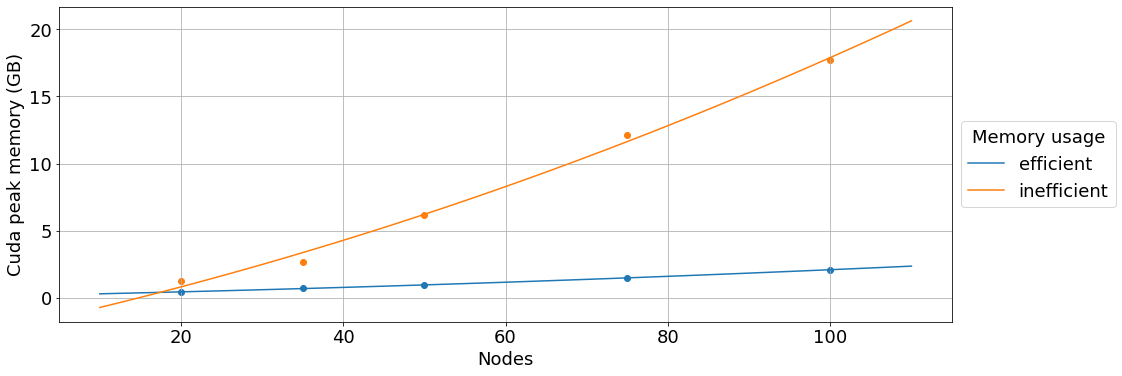}
    \caption{Maximum GPU memory consumption in GB throughout \revised{one} epoch of AM-D training. The line fitted to the memory-inefficient data points is a linear combination of $\{n^{3}, n^{2}, n\}$ and a bias. The line of the memory-efficient data points is a linear combination of $\{n^{2}, n\}$ and a bias.}
    \label{fig:8}
\end{figure*}

\begin{figure*}[t]
    \centering
    \includegraphics[width=\linewidth]{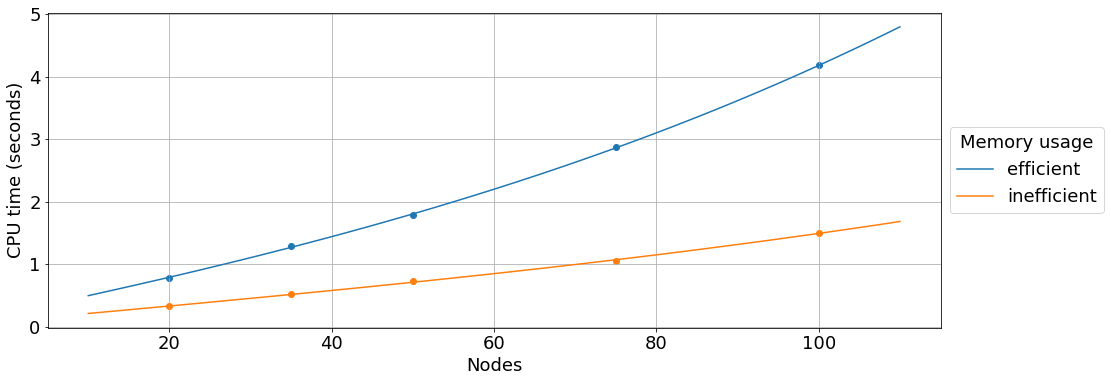}
    \caption{Mean CPU time elapsed across batches of one epoch of AM-D training. The line fitted to both the memory-inefficient and memory-efficient data points is a linear combination of $\{n^{3}, n^{2}, n\}$ and a bias.}
    \label{fig:9}
\end{figure*}

\begin{figure*}[t]
    \centering
    \includegraphics[width=\linewidth]{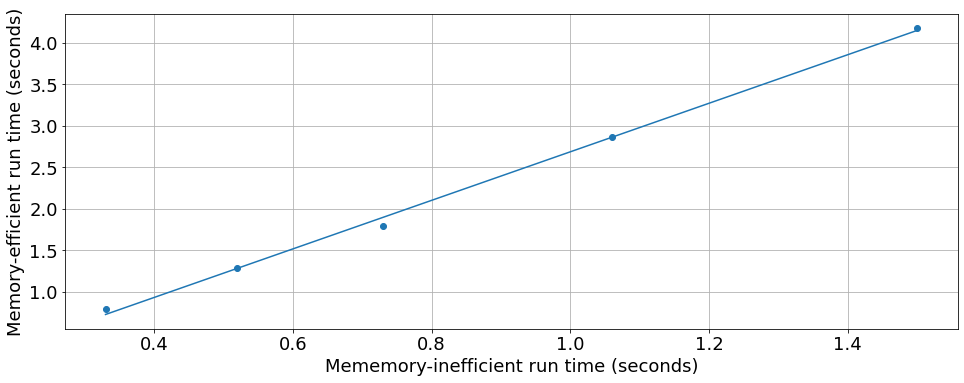}
    \caption{Mean CPU time elapsed across batches of one epoch of AM-D training using the memory-inefficient model vs using the memory-efficient model. The line fitted to the memory-efficient runtime in response to the memory-inefficient runtime has a slope of 2.92 (reflecting the expected 3x increase in runtime).}
    \label{fig:10}
\end{figure*}

The memory consumption results can be seen in Fig.~\ref{fig:8} and the runtime results in Figs.~\ref{fig:9} and~\ref{fig:10}.

As seen in Fig.~\ref{fig:8}, the maximum GPU memory consumption per epoch grows at a significantly lower rate when using the memory-optimized training algorithm. Additionally, as expected, the mean CPU time elapsed per batch is higher when using the memory optimization, as seen in Figs.~\ref{fig:9} and~\ref{fig:10}. We consider this trade-off valuable when \revised{training} the model on CVRPs with several nodes or using memory-restricting GPUs \revised{since} running out of GPU memory crashes the training process.

\clearpage
\clearpage

\section{Conclusion}
\label{section:conclusion}

\revised{In this work, we propose the Combined Deep Constructor and Perturbator~(CDCP) for solving the Capacitated Vehicle Routing Problem. Additionally, we provide a memory improvement in the implementation of the Attention Model-Dynamic. The CDCP algorithm combines an outstanding constructor, the Attention Model-Dynamic, and an exceptional perturbator, the Local Search Heuristic. The CDCP accomplished \revised{promising} results for common testing dataset distributions. Concretely, this method achieved a cost improvement over multiple Deep Learning-based algorithms and showed close results to the state-of-the-art heuristics from the Operations Research field.} In the elaboration of this method, efficiently finding the number of nodes to perturb \revised{at} each iteration step and accurately leveraging Simulated Annealing in the training process of CDCP presented a challenge, as elaborated in the experimental phase.

Conversely, the memory improvement on the existing implementation of the AM-D enables training of this powerful constructor in Capacitated Vehicle Routing Problems with more than 100 nodes. Notably, it allowed us to train this model in our servers with low GPU memory. This was impossible in the past as previous implementations, without the memory improvement, would quickly run out of it. In designing this improvement, obstacles arose while pinpointing code paths with avoidable and high memory consumption. Additionally, leveraging PyTorch's gradient accumulation functionality presented an interesting implementation challenge.

The CDCP has various directions of future work. Exploring a training procedure that simultaneously trains the constructor and perturbator poses an interesting problem. Another compelling challenge would be to try out other combinations of constructors and perturbators. This can extend to intersecting heuristics from the \revised{Operations Research} and \revised{Machine Learning} fields. Finally, expanding CDCP to other combinatorial optimization problems, distinct from the Capacitated Vehicle Routing Problem, seems promising\revised{. The latter could be achieved} by adapting the constructor and perturbator used in this work or using different ones.

\bibliographystyle{ieeetr}
\bibliography{references}

\end{document}